\documentclass[pmlr,twocolumn,10pt]{jmlr} 

 \usepackage{longtable}

\usepackage{booktabs}
\usepackage{siunitx}

\theorembodyfont{\upshape}
\theoremheaderfont{\scshape}
\theorempostheader{:}
\theoremsep{\newline}

\jmlrvolume{LEAVE UNSET}
\jmlryear{2023}
\jmlrsubmitted{LEAVE UNSET}
\jmlrpublished{LEAVE UNSET}
\jmlrworkshop{Machine Learning for Health (ML4H) 2023} 

\title[Knowledge Distillation for Sepsis Outcome Prediction]{A Knowledge Distillation Approach for Sepsis Outcome Prediction from Multivariate Clinical Time Series}

\author{\Name{Anna Wong} \Email{annawong@alum.mit.edu}\\
\Name{Shu Ge} \Email{geshu@mit.edu}\\
\Name{Nassim Oufattole} \Email{nassim@mit.edu}\\
\Name{Adam Dejl} \Email{adamdejl@mit.edu}\\
\Name{Megan Su} \Email{megansu@mit.edu} \\
\Name{Ardavan Saeedi} \Email{av.saeedi@gmail.com}\\
\Name{Li{-}wei H Lehman} \Email{lilehman@mit.edu}\\
\addr Massachusetts Institute of Technology, Cambridge, MA, USA}

\begin{document}

\maketitle

\begin{abstract}
Sepsis is a life-threatening condition triggered by an extreme infection response. Our objective is to forecast sepsis patient outcomes using their medical history and treatments, while learning interpretable state representations to assess patients' risks in developing various adverse outcomes. While neural networks excel in outcome prediction, their limited interpretability remains a key issue.
In this work, we use knowledge distillation via constrained variational inference to distill the knowledge of a powerful "teacher" neural network model with high predictive power to train a "student" latent variable model to learn interpretable hidden state representations to achieve high predictive performance for sepsis outcome prediction.
Using real-world data from the MIMIC-IV database, we trained an LSTM as the "teacher" model to predict mortality for sepsis patients, given information about their recent history of vital signs, lab values and treatments. For our student model, we use an autoregressive hidden Markov model (AR-HMM) to learn interpretable hidden states from patients' clinical time series, and use the posterior distribution of the learned state representations to predict various downstream outcomes, including hospital mortality, pulmonary edema, need for diuretics, dialysis, and mechanical ventilation.
Our results show that our approach successfully incorporates the constraint to achieve high predictive power similar to the teacher model, while maintaining the generative performance.
\end{abstract}

\section{Introduction}
\label{sec:intro}
Sepsis is a life-threatening condition that occurs when the body has an extreme response to an infection. It's estimated that over 1.7 million people in the United States are affected by sepsis each year, and between one-third and one-sixth of those affected will die \citep{CDC} \citep{Evans_2021}. One-third of patients who die in a hospital are affected by sepsis during their stay \citep{CDC}. 

In order to accurately predict a patient's outcome in challenging settings such as during sepsis treatment, we might be interested in training a neural network to make these predictions, and previous approaches have used these models \citep{Bica_2020}, \citep{Li_2021}. However, although neural networks are powerful predictive models, they are not interpretable. 

Latent variable models, such as hidden Markov models (HMM) and other variants of HMMs like the autoregressive HMM (AR-HMM), can be trained on clinical time series data to learn an interpretable latent structure that represents a patient's state of health over time. The limitation of these models is that the state representations are often not optimized for predicting downstream outcomes, and thus the predictions may not be as accurate as with a neural network.

In this paper, we train a model that combines  the neural network approach and the latent variable model approach, with the goal of achieving high predictive power while having an interpretable model. We use knowledge distillation to distill the knowledge learned in a powerful "teacher" model to train a "student" model that has more desirable characteristics such as interpretability.The teacher model is an LSTM with high predictive power, and the student model is an AR-HMM that learns latent states. We use automatic differentiation variational inference (ADVI) for approximate inference to learn the latent states. We then incorporate the teacher model's knowledge through the use of a similarity constraint.

We apply the model on a real-world dataset (MIMIC-IV) \cite{Johnson_2022}.  In addition to evaluating the models' ability to predict mortality, we evaluated the performance of the model in predicting other downstream outcomes such as the development of pulmonary edema, or the need for dialysis, diuretics, or mechanical ventilation.  Our results show that our approach achieves high predictive power similar to the neural network model, while maintaining the generative performance.

\section{Related Works}
\paragraph{Supervised Learning with Probabilistic Graphical Models}A common approach to learning models to predict final outcomes is a 2-stage approach. In the first stage, a graphical model is used to infer latent features for a dataset. These features are then passed into the second stage of the model, which is a discriminative model that predicts an outcome.
Examples of this method include using a graphical model to derive features that are then used as input for a linear regression to predict neuroticism and depression \citep{Resnik_2013}, and using a switching vector autoregressive  model to learn states that are used to predict hospital mortality 
\citep{Lehman_2015}. These types of models often do not have very high predictive power.

\paragraph{Constrained Inference}
Another approach that is used to improve model performance is constrained inference. In this approach, the posterior distribution of learned states is constrained in some way to enforce a performance constraint. For example, we can add a discriminative constraint to improve prediction while still learning interpretable states, by incorporating supervised loss as part of the loss function \citep{Hughes_2018}. 
\citet{Saeedi_2022} used a knowledge distillation approach to improve performance by using a pre-trained discriminative model with high predictive ability to help learn useful features in a more interpretable model. This work builds on \citet{Saeedi_2022} and uses a similarity based constraint.

\section{Methods}
Our approach uses knowledge distillation to develop models that are both predictive and interpretable. Knowledge distillation is a technique that uses a model with high predictive power, known as the teacher model, and distills its information into another model known as the student model, which ideally has some preferable characteristics such as interpretability. For our work, we first trained the teacher model, which was an LSTM, to learn patient hospital mortality. For the student model, we chose to use an autoregressive HMM (AR-HMM) to learn hidden states. To learn this student model, we used a variational inference technique known as ADVI to learn the latent states of the model. While learning the AR-HMM, we trained a recognition network to learn latent states of the model while also incorporating a similarity-based constraint between the teacher model and the student model. The similarity-based constraint was used to distill knowledge from the teacher to the student.
\subsection{Variational Inference}
The mechanism behind many models that identify patterns in a model or make predictions is typically a computation of the posterior distribution of latent variables, given the observation, $p(z, \theta|x)$, where $z$ is local latent variables, $\theta$ is global latent variables, and $x$ is observations. However, it is often difficult to calculate the posterior, which leads to the use of approximate inference techniques like variational inference. A typical framework of variational inference maximizes the Evidence Lower Bound (ELBO) with respect to the variational parameters $\phi = \{\phi_\theta, \phi_z\}:$

\begin{flalign}
    \mathcal{L}(\phi_\theta, \phi_z;x)
    \triangleq \mathbb{E}_{q\phi}[\log p(x,z,\theta)
    &-\log q_{\phi_\theta}(\theta)\\
    &-\log q_{\phi_z}(z|x)],
\end{flalign}
which is a lower bound on the log-likelihood. To make our approach more accessible to a wider range of applications, our framework, similar to \citet{Saeedi_2022}, uses Automatic Differentiation Variational Inference (ADVI) to approximate our global latent variables $\theta$ and a recognition network for our local latent variables $z$.

\paragraph{Global Latent Variables $\theta$}
ADVI is an example of one such technique. It is a flexible black-box variational inference method that can be used for many different probabilistic models. It achieves this by transforming the $K$-dimensional latent variables $\theta$ such that they live in the real coordinate
space, $\mathbb{R}^K:T:\text{supp}(p(\theta))\rightarrow \mathbb{R}^K$, so
ADVI can choose the variational distribution independent of
the generative model. Note also the variational approximation in the original space can then be written as $q(\theta;\phi_\theta) = q(T(\theta); \phi_\theta)|\text{det}(J_T(\theta))|$.
Then, one can further assume a factorized
Gaussian distribution is the variational approximation for
the transformed latent variables: $q(T(\theta); \phi_{T(\theta)})=\mathcal{N}(T(\theta); \mathbf{\mu}, \text{diag}(\exp(\mathbf{\omega})^2))$, where $\phi_{T(\theta)} = (\mu_1, \dots, \mu_K, \omega_1, \dots, \omega_K)$ are the variational parameters in the unconstrained space. Note that these
implicitly induce non-Gaussian variational distributions in the original latent variable space \citep{kucukelbir2016automatic}.  

\paragraph{Local Latent Variables $z$}
On the other hand, a recognition network is used to map from observations to a posterior distribution of local latent variables. They can be used to efficiently approximate parameters of latent variables \citep{kingma2023variational}. The recognition network maps observations $x$ into the approximate posterior $q(z|x)$, which is typically used as the features that are inputted to a predictive task. Similar to the ADVI case, we transform the local latent variables into real coordinate space: $\zeta = T(z)$ and assume a factorized Gaussian distribution 
$q_{\phi_z}(\zeta|x) = \mathcal{N}(\mu_{\phi_z}(x), \text{diag}(\exp(\omega_{\phi_z}(x))^2))$, where $\phi_z$ is the parameter set of the recognition network. In Saeedi et al., the authors use Autoencoding Variational Bayes (AEVB) to train a recognition network that is used to infer local latent variables of the HMM \citep{Saeedi_2022}. 

\paragraph{ELBO}
The outputs of the recognition network and ADVI are used to get posterior samplers of local and global latent variables to compute ELBO, 
\begin{flalign*}
    \mathcal{L}(\phi_\theta, \phi_z; x_i)
    &\triangleq \mathbb{E}_{q_\phi}[
    \log p(x_i, T^{-1}(\Theta), T^{-1}(\zeta))\\
    & + \log |\text{det}(J_{T^{-1}}(\Theta))|\\
    & + \log |\text{det}(J_{T^{-1}}(\zeta))|]\\
    & + \mathbb{H}(q_{\phi_\theta}(\Theta))\\ & + \mathbb{H}(q_{\phi_z}(\zeta|x_i)).
\end{flalign*}
using Monte Carlo approximation. Using the parameterization trick in \citet{kucukelbir2016automatic}, we can rewrite the expectation in standard Gaussian density:
\begin{flalign*}
    \mathcal{L}(\phi_\theta, \phi_z; x_i)
    &\triangleq \mathbb{E}_{\mathcal{N}(\epsilon;o,I)}[
    \log p(x_i, T^{-1}(\Theta_\epsilon), T^{-1}(\zeta_\epsilon))\\
    & + \log |\text{det}(J_{T^{-1}}(\Theta_\epsilon))|\\
    & + \log |\text{det}(J_{T^{-1}}(\zeta_\epsilon))|]\\
    & + \mathbb{H}(q_{\phi_\theta}(\Theta))\\ & + \mathbb{H}(q_{\phi_z}(\zeta|x_i)).
\end{flalign*}

\paragraph{Similarity Constraint}
\label{sec:knowledge-distillation-constraint}
We incorporated the knowledge distillation constraint by using a similarity-based constraint between the teacher and student models. We used the constraint presented in \citet{Saeedi_2022}. Let $C_{t}$ be the feature dimensionality of the teacher model, and $C_{s}$ be the feature dimensionality of the student model. For a dataset of size $N$, we denote the feature representations of the teacher and student models as $F^{t} \in \mathbb{R}^{N \times C_{t}}$ and $F^{s} \in \mathbb{R}^{N \times C_{s}}$, respectively. For the student model, we assume that every row of the feature representation is a function of the inferred latent variables. The knowledge distillation constraint is designed to make sure that the differences between the two feature representations is less than some tolerance level.

More specifically, we compute the similarity of feature representations across patients by taking the dot product, which results in $N \times N$ matrices:
\begin{equation}
    \tilde{F}^{s} = F^{s} \cdot F^{s\top} \text{ and } \tilde{F}^{t} = F^{t} \cdot F^{t\top}
\end{equation}

We also apply a normalization to the matrices. After taking the dot product of the feature matrix, we normalize by dividing each column by the $\ell^{2}$ norm of the row. Let us denote the final normalized similarity matrices as $\bar{F}^{s}$ and $\bar{F}^{t}$ for the student and teacher models, respectively. We calculate the similarity loss as 

\begin{equation}
    \text{similarity loss} = \gamma\frac{1}{N^{2}} ||\bar{F}^{s} - \bar{F}^{t}||^{2}
\end{equation}
where $\gamma$ is a hyperparameter that specifies how much to weight the loss from this similarity constraint in the overall loss function.

The final objective function is 
\begin{equation}
    \min_{\phi_\theta, \phi_z} -\mathcal{L}(\phi_\theta, \phi_z; x_i) + \text{similarity loss},
\end{equation}
where the variational objective is regulated by the knowledge distillation constraint so that we maximize ELBO while ensuring that the student model has similar features as the teacher model. Then a gradient descent method is used to update global latent variables and parameters of the recognition networks such that they simultaneously maximize ELBO.
The inferred posterior of local latent variables are used to perform downstream predictions, including posterior topic proportions in a topic model or marginal
posterior of latent states in a HMM model \citep{Saeedi_2022}.

\subsection{AR-HMM}
In an AR-HMM, each state is parameterized by a set of AR coefficients (which define how much the previous observations affect the current observation), a covariance matrix $\Sigma$, and  a bias term $b$,  so we can define the $k$-th state as $\theta_{k} = \{ A_{k}, \Sigma_{k}, b_{k} \}$. In an AR-HMM process with order $r$, each observation depends on the $r$ observations before it \citep{Fox_2010}. Let $x_{t}^{(i)}$ be the observation vector of the $i$-th patient at time $t$, and let $z_{t}^{(i)}$ be the state of the corresponding Markov chain for that patient at time $t$. Let $\pi_{k}$ be the transition probabilities for state $k$. Then, since this is a Markov chain, we know that $z_{t}^{(i)} \sim \pi_{z_{t-1}^{(i)}}$, for all $t > 1$. An order $r$ AR-HMM process, denoted by VAR($r$), is defined as follows
\begin{flalign}\label{eq:arhmm}
&z_{t}^{(i)} \sim \pi_{z^{(i)}_{t-1}}\\
&x_{t}^{(i)} = \sum^{r}_{l=1} A^{z_{t}^{(i)}}_{l} x_{t-l}^{(i)} + e_{t}^{(i)}(z_{t}^{(i)}) + b_{z_{t}^{(i)}} \\
&\triangleq A_{z_{t}^{(i)}} \tilde{x}_{t}^{(i)}  + e_{t}^{(i)}(z_{t}^{(i)}) + b_{z_{t}^{(i)}}
\end{flalign}
where $e_{t}^{(i)}(z_{t}^{(i)}) \sim \mathcal{N}(0, \Sigma_{(Z_{t})})$ is the state-specific noise, $A_{k} = [A^k_1 ... A^k_r]$ are the lag matrices that indicate how much to weight previous observations, and $\tilde{x}_{t}^{(i)} = [x_{t-1}^{(i)^{\top}} ... x_{t-r}^{(i)^{\top}}]^{\top}$ are the previous observations. 

\subsection{Data} In this work, we used data from the Medical Information Mart for Intensive Care IV (MIMIC-IV) database \citep{Johnson_2022}. For our patient cohort, we selected patients meeting the sepsis-3 criteria. After ensuring patients met all inclucsion and exclusion criteria (see Appendix), we were left with 7,663 patients. Hospital mortality of the cohort is approximately 13\%. We put 70\% of patients in the training set, and 15\% in each of the validation and testing sets. A detailed description of our cohort and a full list of covariates are provided in Appendix \ref{apd:data}.

\subsection{Teacher Model}
For our teacher model, we trained an LSTM on the patient data to predict mortality. The teacher model included a total of 47 features, which is more features than the student model. The extra features included in the LSTM but not the student model are shown in Table \ref{tab:lstm_time_varying_vars}. 
Note that the teacher model and the student model need not have the same input dimension.  In our experimental setup, the teacher LSTM is fed with higher dimensional time series data as input to predict downstream outcomes.  Our premise was that the knowledge learned by the teacher model could be transferred to the student model, without the student model needing as many covariates. 
We tried various seeds and hyperparameter settings for this LSTM, and used the model with the best validation AUROC for our teacher model.

\subsection{Student Model}
For our student models, we used an autoregressive hidden Markov
Model (AR-HMM) of order 1, with $D$-dimensional Gaussian
distribution, where $D$ is the number of covariates used for each patient's input to the model. In our models, $D = 34$. For each patient, we have a $D \times T$ vector input, where $T = 24$ is the number of timesteps. There are $K$ possible latent states learned by the AR-HMM.The models we discuss in this paper use an AR-HMM with AR order 1.

The baseline model was a basic AR-HMM without knowledge distillation or supervision, run for 20 iterations. The model that incorporates knowledge distillation via a similarity constraint is denoted as KD-AR-HMM. We also used a model with a discriminator constraint (DISC-AR-HMM). The models with constraints were run for 10 iterations.

\subsection{Outcome Prediction}
For the baseline AR-HMM and the KD-AR-HMM, we used a logistic regression model that took as input the features outputted by the AR-HMM model, and outputted the probability of an outcome such as mortality. For the models with the discriminator constraint, we used the trained discriminator network to make predictions. In addition to predicting mortality, we also used the same set of features to predict other patient outcomes such as the development of pulmonary edema, or the need for dialysis, mechanical ventilation, or diuretics. We adjusted various hyperparameters in order to tune the model. A description of our hyperparameter search can be found in Appendix \ref{apd:hyperparams}.

\section{Results}
\subsection{Mortality Prediction}

We adapted the existing AR-HMM models from \citet{Saeedi_2022} to apply them to the MIMIC-IV dataset to predict patient mortality. The results presented in this paper have the number of states as $K = 5$. We also tried using 10, 15, and 20 states, and the KD-AR-HMM achieved similar performance as the 5 state version in \tableref{tab:model-results}, and also consistently outperformed other baselines. We ran each model with 10 different seeds and hyperparameter settings, and chose the best models based on the validation AUC. The baseline had a low AUROC of 0.540. The model with the discriminator constraint performed better than the baseline with an AUROC of 0.633. The model that incorporated knowledge distillation performed the best, with an AUROC of 0.796.

\begin{table}[htbp]
\floatconts
  {tab:model-results}
  {\caption{Performance in mortality prediction.}}
  {\begin{tabular}{lrr}
  \bfseries Model & \bfseries AUC & \bfseries Log Likelihood \\ \hline
    LSTM (Teacher) & 0.833 & N/A \\
    AR-HMM (Baseline) & 0.540 & -1.73E+06  \\
    DISC-AR-HMM & 0.633 & -4.31E+07  \\
   \textbf {KD-AR-HMM} & \textbf{0.796} & -2.73E+06  \\
  \end{tabular}}
\end{table}

In addition to measuring the AUROC to evaluate the predictive ability of our models, we measured the log likelihood to evaluate the generative ability of the models. Compared to the baseline AR-HMM, the constrained models had comparable log likelihoods, indicating that the constraints did not significantly decrease the model fit for the data. In particular, the log likelihood of the KD-AR-HMM was the closest to the baseline. A summary of these metrics can be found in \tableref{tab:model-results}.

\subsection{Other Outcome Predictions}

\begin{table*}[htbp!]
\floatconts
  {tab:other_outcomes}
  {\caption{Performance for Other Outcome Predictions in AUROC. MV = Mechanical Ventilation}}
  {\begin{tabular}{lrrrrr}
  \bfseries  & \bfseries Mortality & \bfseries Edema & \bfseries Dialysis & \bfseries MV & \bfseries Diuretic\\ \hline
    LSTM (Teacher) & 0.833 & 0.663 & 0.817 & 0.601 &  0.640\\ \hline
    AR-HMM & 0.540 & 0.481 & 0.563 & 0.453 & 0.488 \\
    DISC-AR-HMM & 0.633 & 0.584 & 0.805 & 0.489 & \textbf{0.583} \\
    KD-AR-HMM  & \textbf{0.796} & \textbf{0.589} & \textbf{0.819} & \textbf{0.638} & 0.563 \\
  \end{tabular}}
\end{table*}

In this section, we assess our model's ability to predict other downstream clinical outcomes,  including edema, need for dialysis, diuretics, and mechanical ventilation in the next 48 hours (i.e. within the first 72 hours of the patients' ICU stays).  For each outcome, we took the existing dataset and removed any patients who had that particular outcome within the first 24 hours.

For the AR-HMM based models, we used the same set of features (state marginals) to train logistic regression models for each of the clinical outcomes.  Specifically, note that for the KD-AR-HMM model, we used the same state marginals learned through knowledge distillation from the Teacher LSTM hospital mortality model as input to build a classifier in predicting each respective clinical outcome. As such, the KD-AR-HMM model in these experiments did not have access to the labeling information from these other clinical outcomes during training and inference. 

Results from the individual models trained to predict each outcome are summarized in \tableref{tab:other_outcomes}. 
We show the test set AUROCs of the LSTM models trained to predict each of the respective outcomes; they serve as a reference for AUROCs achieved from direct supervised learning using LSTM. 
For the baseline AR-HMM and KD-AR-HMM models, as well as for the non-mortality outcomes for the DISC models, we used the 2-stage process of fitting a logistic regression to predict the outcome. For the mortality outcomes for the DISC models, we used a 1-stage approach of using the discriminator network to make predictions. KD-AR-HMM in general achieved better predictive performance than the AR-HMM and DISC-AR-HMM baselines, except in the case of predicting need of diuretics.
This demonstrates that KD-AR-HMM is able to effectively learn representations that are useful in predicting multiple downstream tasks.

\paragraph{Similarity Matrix}

\begin{figure}[htbp]
\subfigure[LSTM]{\label{fig:sim-matrix-lstm}%
\includegraphics[width=0.45\linewidth]{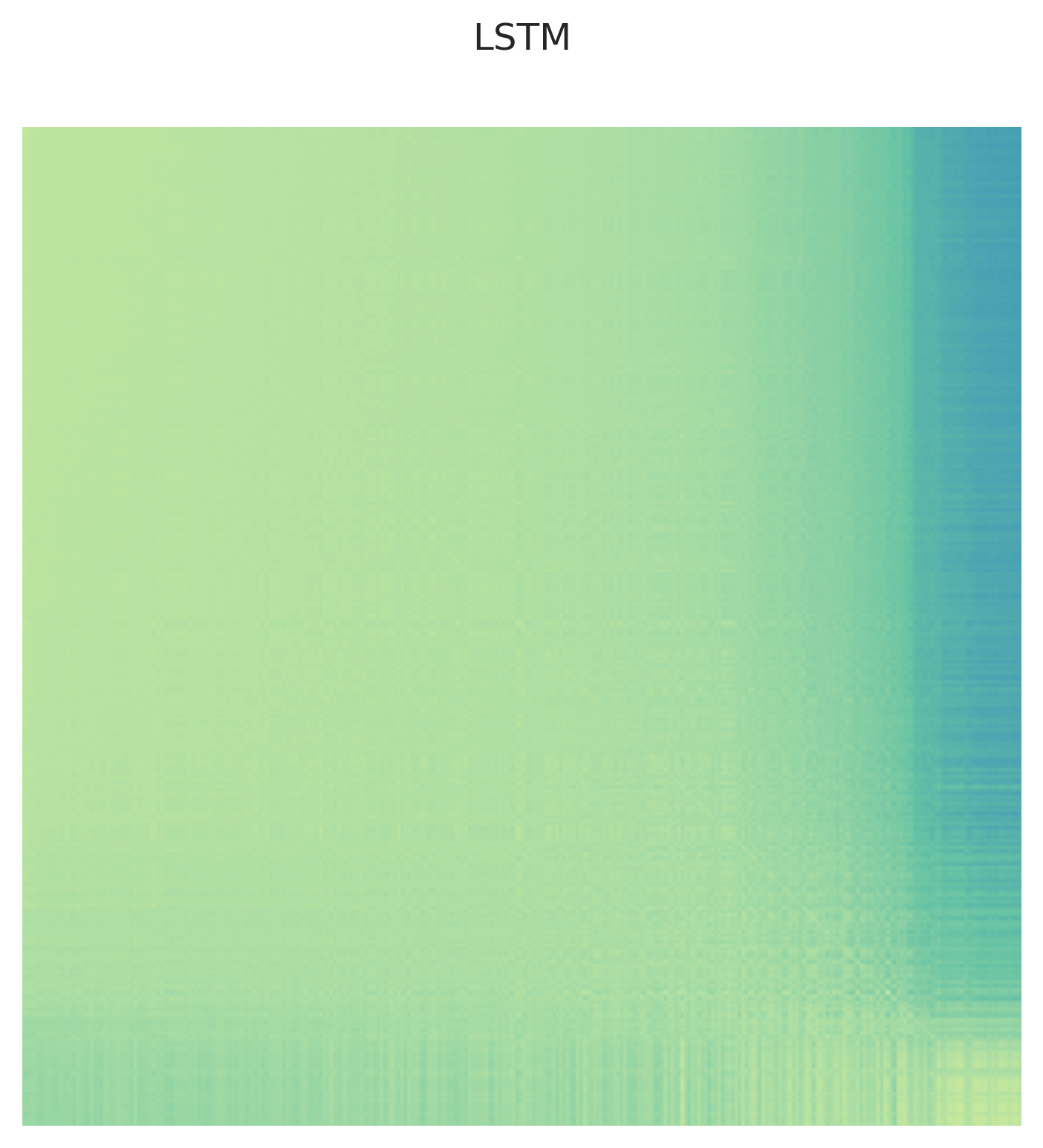}} \quad
\subfigure[KD-AR-HMM]{\label{fig:sim-matrix-kd}
\includegraphics[width=0.45\linewidth]{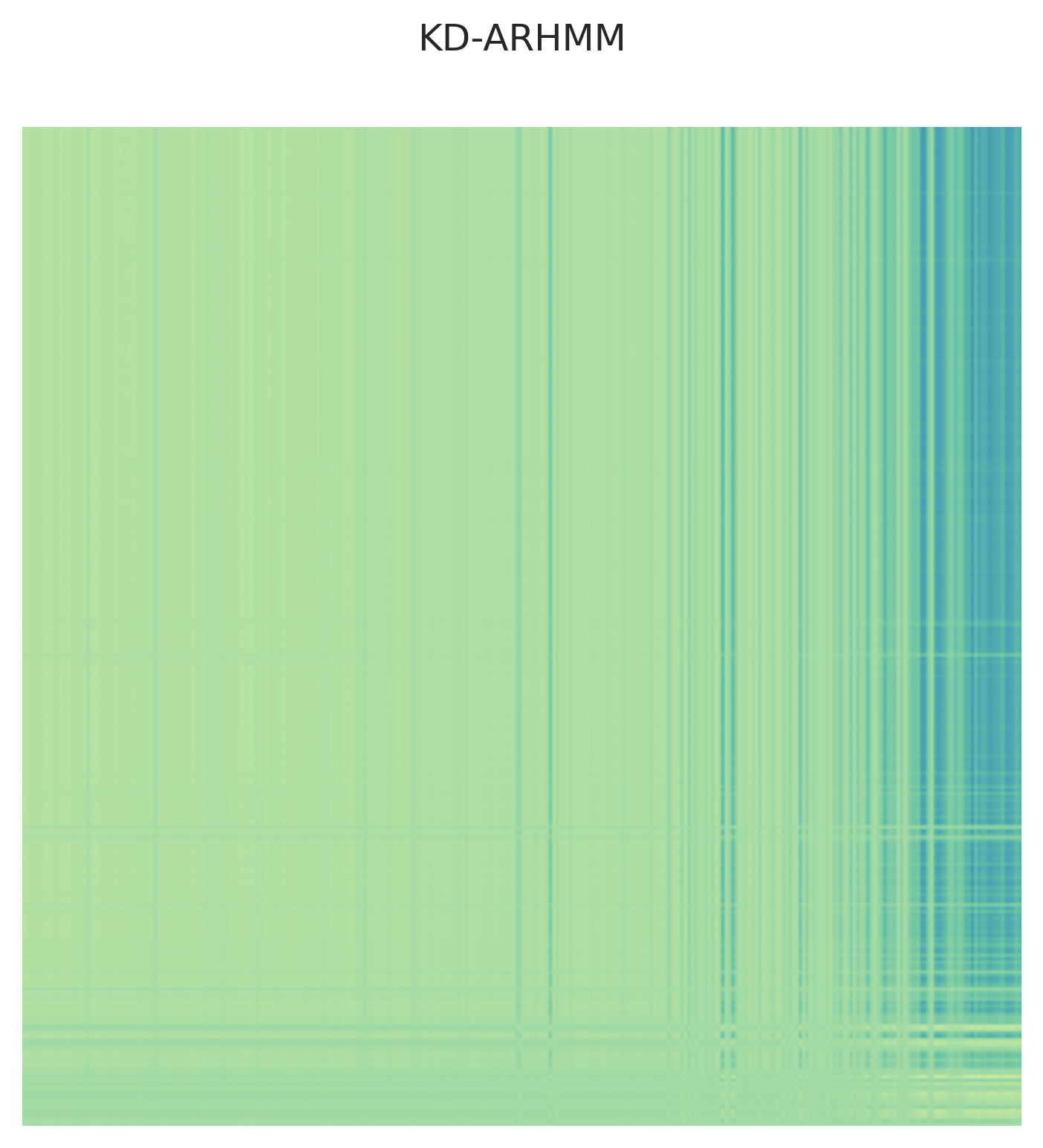}}
\caption{\small{Pairwise similarity matrices for the teacher model (LSTM) and the KD-AR-HMM. Brighter colors indicate higher similarity between patients.}}
\label{fig:sim_matrices}
\end{figure}

Figure~\ref{fig:sim_matrices} shows pairwise similarity matrices for the test set for the teacher
LSTM model, and the KD-AR-HMM that is constrained to be similar to the teacher model.  Each row and column corresponds to a patient in the test dataset. The constrained model is similar to the teacher model, indicating success with the knowledge distillation technique.

\paragraph{Association Analyses of Learned States and Outcomes.} We conducted a logistic regression analysis to identify states that were significantly associated with the outcomes (see Appendix for details). We found that, for KD-AR-HMM, state 1 was a low-risk state, as increasing proportion of it was significantly  associated with a lower odds in mortality, edema, and dialysis. State 3 was a high-risk state, as increasing proportion of it was significantly associated with higher odds in mortality, edema, and dialysis.

\paragraph{Individual Patient Analyses}
\begin{figure}[htbp]
\floatconts
  {fig:patient-ex}
  {\caption{\small{Example trajectories of a patient who lived (top) and a patient who died (bottom).}}}
  {\includegraphics[width=1\linewidth, height=1\linewidth]{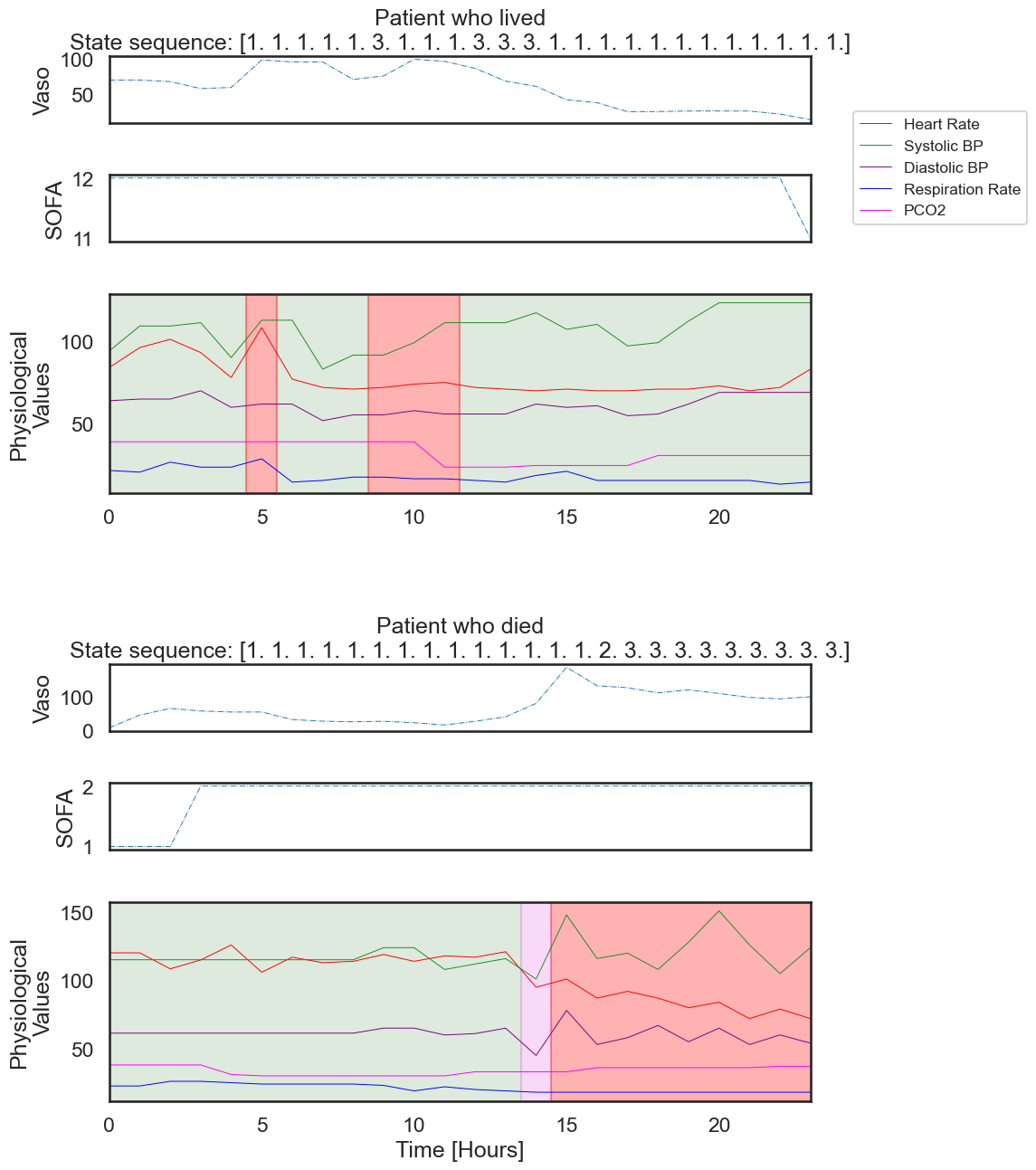}}
\end{figure}
Figure \ref{fig:patient-ex} shows an example of a patient who lived and a patient who died, and their states throughout the first 24 hours. The patient who lived had a few hours of being in the high-risk state (state 3, in red) in the first half of their stay, but then went back to the low-risk state (state 1, green). We can also see that the patient's return to the low-risk state seems to coincide with a decrease in vasopressor fluids administered. On the other hand, the patient who died started off in the low-risk state, but towards the end of the 24 hours, was in the high risk state. We can also see that this patient had an increase in the vasopressor fluids and SOFA score.



\section{Discussions and Conclusion}

In this work, we  use knowledge distillation to distill the high predictive power of a neural network model to a latent variable model to learn   state representations from multivariate clinical time series data for sepsis outcome prediction.  Experimental results using the MIMIC-IV database show that, our approach successfully incorporates the constraint to achieve high predictive power similar to the neural network model, while being able to learn interpretable state representations to assess patient's risks in developing various downstream adverse outcomes.  

\section{Acknowledgements}
L Lehman was in part funded by MIT-IBM Watson AI Lab and NIH grant R01EB030362.

\bibliography{jmlr-sample}

\appendix

\section{Data and Cohort}
In this work, we used data from the Medical Information Mart for Intensive Care IV (MIMIC-IV) database, which contains medical records from hospital admissions and ICU stays at the Beth Israel Deaconess Medical Center (BIDMC) \citep{Johnson_2022}.  
For our patient cohort, we selected patients meeting the sepsis-3 criteria. Under this criteria, a patient is defined to have sepsis if they have both an episode of suspected infection and a Sequential Organ Failure Assessment (SOFA) score of 2 or more points. An episode of suspected infection is defined as either (a) an antibiotic was given and a culture was sampled within 24 hours or (b) a culture was sampled and an antibiotic was administered within 72 hours.

The dataset excludes patients whose time of suspected infection was more than 24 hours after ICU admission. It also excludes patients who were admitted after cardiac, vascular, or trauma surgery since those surgeries pose risks that could lead to different mortality outcomes. Additionally, if a patient had more than one ICU stay, only the first stay was used. The dataset also excludes patients who did not have documented pre-ICU fluids.
Finally, we removed patients who died within 24 hours of entering the ICU, and patients who did not have all 24 hours of data.

After ensuring patients met all of these criteria, we were left with 7,663 patients. We put 70\% of patients in the training set, and 15\% in each of the validation and testing sets. The mean age of patients was 65.10, the median age was 67.0, and 4135 of the patients were male.

For each patient, we have 24 hours of hourly data, with covariates such as heart rate, blood pressure, SOFA score, and other clinical variables such as glucose, creatinine, potassium, and more. We also have information about the actual treatment given to these patients at each hour, such as if patients are on mechanical ventilation or dialysis and the amount and dosage (if any) of fluids, vasopressors, and diuretics given. Finally, we have information about outcomes such as if the patient has pulmonary edema, is on diuretics, dialysis, or mechanical ventilation, or if they die in the hospital. We fill in missing covariate values by either extending the previous covariate measurement for that patient if it exists, or filling it in with the population average value for that covariate.

\paragraph{Other Clinical Outcomes.} The final datasets had the following total number of remaining cohort, and percentage of patients with the adverse outcomes within the cohort (i.e. patients who ultimately experienced that outcome):  4,784 (18.1\%) for edema, 7,103 (2.2\%) for dialysis, 3,965 (5.3\%) for mechanical ventilation, and 6,108 (15.1\%) for diuretics. 

\subsubsection{Association Analysis}
To determine whether or not a state is significantly associated with mortality, we looked at the odds ratios and p-values for each state. We fitted a logistic regression on the test set, where the features were the features generated by the KD-AR-HMM and the labels were the mortality outcomes.

We used univariate logistic regressions to try and isolate the effect of each individual state on the outcome. We used two methods of creating features for input to the regression - one method used the state marginals, while the other used the Viterbi algorithm to calculate the patient's probability of being in a state.  We adjusted the p-values using an FDR (False Discovery Rate) adjustment to account for the fact that we fitted several distinct univariate logistic regressions. 

For the univariate association analyses with respect to each clinical outcome, we use the Viterbi algorithm to derive the most likely state sequences and use the probability distributions over each individual state  as input to the univariate logistic regression model.  Table~\ref{tab:odds_ratio} shows the results from the univariate analysis using Viterbi-derived state distributions for the KD-AR-HMM; results show that states 1, 2 and 3 are significantly associated with the mortality outcomes. We show the three states with the smallest p-values. 

\begin{table}[htbp]
\label{tab:odds_ratio}
\floatconts
  {tab:odds_ratio}
  {\caption{Hospital Mortality: Univariate odds ratio (OR) for the KD-AR-HMM model, using states predicted by the Viterbi algorithm, with features multiplied by 100.}}
  {\begin{tabular}{lrrrrr}
  \bfseries State & \bfseries OR & \bfseries adj p-value \\ \hline
    1 & 0.947 (0.932, 0.961) & \bfseries \textless 0.001 \\
    2 & 1.087 (1.014, 1.164) & \bfseries 0.031 \\
    3 & 1.057 (1.039, 1.075) & \bfseries \textless 0.001 \\
  \end{tabular}}
\end{table}

We also used multivariate logistic regressions to calculate the odds ratios. Table~\ref{tab:odds_ratio_multi}, shows the results from the multivariate analysis using Viterbi-derived state distributions for the KD-AR-HMM; results show that states 1 and 3 are significantly associated with the mortality outcomes, which matches the univariate analysis.

\begin{table}[htbp]
\label{tab:odds_ratio_multi}
\floatconts
  {tab:odds_ratio_multi}
  {\caption{Hospital Mortality: Multivariate odds ratio (OR) for the KD-AR-HMM model, using states predicted by the Viterbi algorithm, with features multiplied by 100.}}%
  {\begin{tabular}{lrrrrr}
  \bfseries State & \bfseries OR & \bfseries adj p-value \\ \hline
    1 & 0.9800 (0.9781, 0.9819) &  \bfseries \textless 0.001 \\
    2 & 1.0459 (0.9818, 1.1141) & 0.164 \\
    3 & 1.0311 (1.0149, 1.0476) & \bfseries \textless 0.001 \\
  \end{tabular}}
\end{table}

We conducted the same odds ratio analyses for the dialysis outcome. Table~\ref{tab:odds_ratio_dialysis}, shows the results from the univariate analysis of dialysis outcomes using Viterbi-derived state distributions for the KD-AR-HMM; results show that states 1 and 3 are significantly associated with the dialysis outcomes.

\begin{table}[htbp]
\label{tab:odds_ratio_dialysis}
\floatconts
  {tab:odds_ratio_dialysis}
  {\caption{Dialysis: Univariate odds ratio (OR) for the KD-AR-HMM model for dialysis outcome, using states predicted by the Viterbi algorithm, with features multiplied by 1000.}}%
  {\begin{tabular}{lrrrrr}
  \bfseries State & \bfseries OR & \bfseries adj p-value \\ \hline
    1 & 0.997 (0.995, 0.999) & \bfseries 0.010 \\
    2 & 1.010 (0.997, 1.022) & 0.187 \\
    3 & 1.003 (1.001, 1.005) & \bfseries 0.010 \\
  \end{tabular}}
\end{table}

We also used multivariate logistic regressions to calculate the odds ratios. Table~\ref{tab:odds_ratio_dialysis_multi}, shows the results from the multivariate analysis using Viterbi-derived state distributions for the KD-AR-HMM; results show that state 1 is significantly associated with the dialysis outcomes, which partially matches the univariate analysis.

\begin{table}[htbp]
\label{tab:odds_ratio_dialysis_multi}
\floatconts
  {tab:odds_ratio_dialysis_multi}
  {\caption{Dialysis: Multivariate odds ratio (OR) for the KD-AR-HMM model for dialysis outcome, using states predicted by the Viterbi algorithm, with features multiplied by 100.}}%
  {\begin{tabular}{lrrrrr}
  \bfseries State & \bfseries OR & \bfseries adj p-value \\ \hline
    1 & 0.965 (0.961, 0.969) &  \bfseries \textless 0.001 \\
    2 & 1.036 (0.914, 1.174) & 0.580 \\
    3 & 0.992 (0.974, 1.011) & 0.400 \\
  \end{tabular}}
\end{table}

\section{Hyperparameter Tuning}\label{apd:hyperparams}
In order to tune the model, we considered adjusting various hyperparameters. The objective function used to train the AR-HMMs was made up of several different components, so we tried different coefficients for each component. If the coefficient was too low, then that loss component would have little effect on the model's predictions. However, if the coefficient was too high, that loss term would dominate the overall loss, and the network might fail to learn the true state marginals. The options we considered for each of these hyperparameters is shown in Table \ref{tab:hyperparams}.

\begin{table}[htbp]
\floatconts
  {tab:hyperparams}%
  {\caption{Hyperparameter Settings. This table shows the options for various hyperparameters that we tried for our AR-HMM models.}}%
  {\begin{tabular}{ll}
  \bfseries Hyperparameter & \bfseries Settings\\ \hline
    Discriminator coefficient & 1e5, 1e7, 1e9 \\
    Similarity coefficient & 1e9, 1e11, 1e13, 1e15 \\
    Log-likelihood coefficient	   & 1, 1e3, 1e6 \\
    determinants coeff	   & 1, 1e3, 1e6, 1e9 \\
    global vars entropy coeff	   & 1, 1e3, 1e6 \\
    local vars entropy coeff	   & 1, 1e3, 1e6 \\
    Priors coefficient	   & 1, 1e5, 1e10 
  \end{tabular}}
\end{table}
    
A grid search through all of these combinations was too computationally expensive. Thus, for each seed, we randomly selected values from the search space.

\section{Data Covariates}\label{apd:data}
Table \ref{tab:time_varying_vars} shows the variables that were used as inputs for both the teacher LSTM and student AR-HMM models. The vasopressor amount variable measures the total amount of vasopressors used during that time period. Vasopressors are standardized by comparing their relative strength to norepinephrine, also known as noradrenaline or norad. The vasopressors included in this standardization are norepinephrine (or levophed), epinephrine, vasopressin, phenylephrine, and dopamine. The measurement unit used is mcg/kg/minute, except for vasopressin, which is expressed as units/minute. The standardization process involves adjusting the dosage rate of each vasopressor by multiplying it with a scaling constant based on the typical dosing of each drug. Norepinephrine is typically administered at a dosage range of 0-1 mcg/kg/minute. If multiple vasopressors were used during the same time period, the combined total dose for each hour is reported.

\begin{table*}[htbp]
\floatconts
{tab:time_varying_vars}
{\caption{MIMIC time-varying variables that were used as inputs to both the teacher LSTM and student AR-HMM models.}}
{
    \begin{tabular}{ccc} 
    \bfseries Variable Name & \bfseries Variable Type & \bfseries Units \\ \hline
     Heart Rate & Continuous & beats/min \\
     Diastolic Blood Pressure & Continuous & mmHg \\
     Systolic Blood Pressure & Continuous & mmHg \\
     Mean Blood Pressure & Continuous & mmHg \\
     Minimum Diastolic Blood Pressure & Continuous & mmHg \\
     Minimum Systolic Blood Pressure & Continuous & mmHg \\
     Minimum Mean Blood Pressure & Continuous & mmHg \\
     Temperature & Continuous & \textdegree C \\
     SOFA Score & Treated as Continuous & N/A \\
     Glasgow Coma Score & Treated as Continuous & N/A \\
     Platelet & Continuous & counts/$10^9$L \\
     Hemoglobin & Continuous & g/dL \\
     Calcium & Continuous & mg/dL \\
     BUN & Continuous & mmol/L \\
     Creatinine & Continuous & mg/dL \\
     Bicarbonate & Continuous & mmol/L \\
     Lactate & Continuous & mmol/L \\
     Potassium & Continuous & mmol/L \\
     Bilirubin & Continuous & mg/dL \\ 
     Glucose & Continuous & mg/dL \\ 
     pO2 & Continuous & mmHg \\
     SO2 & Continuous & \% \\
     SpO2 & Continuous & \% \\
     pCO2 & Continuous & mmHg \\
     Total CO2 & Continuous & mEq/L \\
     pH & Continuous & Numerical[1,14] \\
     Base excess & Continuous & mmol/L \\
     Weight & Continuous & kgs \\
     Respiratory Rate & Continuous & breaths/min \\
    
     Total Fluids & Continuous & mL \\
     
     Urine Output & Continuous & mL \\
     Total Urine Output & Continuous & mL \\
     Fluid Bolus & Continuous & mL \\ 
     Vasopressor Amount & Continuous & mcg/kg/min \\ 
    \end{tabular}
}
\end{table*}

\begin{table*}[htbp]
\floatconts
  {tab:lstm_time_varying_vars}%
  {\caption{MIMIC time-varying variables that were only used for the teacher LSTM model, and not for the student AR-HMM models.}}%
    {
    \begin{tabular}{ccc} 
     \bfseries Variable Name & \bfseries Type & \bfseries Units \\  \hline
     Minimum Mean Blood Pressure from Baseline & Continuous & mmHg \\
     Glasgow Coma Score - Motor & Ordinal& N/A \\
     Glasgow Coma Score - Verbal & Ordinal & N/A \\
     Glasgow Coma Score - Eye & Ordinal & N/A \\
     O2 requirement level & Ordinal [0,6] & N/A\\
     Pulmonary Edema Indicator & Binary &N/A \\
     Cumulative Edema & Binary &N/A \\
     Diuretics Indicator & Binary &N/A \\
     Diuretics Amount & Continuous & mg \\ 
     Dialysis Indicator & Binary &N/A \\
     Mechanical Ventilation Indicator & Binary &N/A \\
     Bolus Indicator & Binary &N/A \\
     Vasopressor Indicator & Binary &N/A \\
    \end{tabular}
}
\end{table*}

\end{document}